\documentclass[conference]{IEEEtran}
\IEEEoverridecommandlockouts
\usepackage[top=1in,left=0.75in,right=0.75in,bottom=0.75in]{geometry}
\usepackage{afterpage}
\usepackage{times}

\usepackage{cite}
\usepackage{multicol}
\usepackage{amsmath}
\usepackage{graphicx}
\usepackage{gensymb}
\usepackage{amsmath, xparse}
\usepackage{mathrsfs}
\usepackage{soul}
\usepackage{float}
\usepackage{xcolor}
\usepackage{stfloats}
\usepackage[bookmarks=true]{hyperref}

\def\BibTeX{{\rm B\kern-.05em{\sc i\kern-.025em b}\kern-.08em
    T\kern-.1667em\lower.7ex\hbox{E}\kern-.125emX}}

\def\BibTeX{{\rm B\kern-.05em{\sc i\kern-.025em b}\kern-.08em
    T\kern-.1667em\lower.7ex\hbox{E}\kern-.125emX}}

\def\ie{{\em i.e.,\ }}

\begin{document}

\title{Bio-inspired tail oscillation enables fast crawling on deformable granular terrains} 
\IEEEoverridecommandlockouts


\author{Shipeng Liu$^{1}$, Meghana Sagare$^{1}$, Shubham Patil$^{1}$, and Feifei Qian$^{*,1}$\\

\thanks{$^*$Corresponding author: Feifei Qian (feifeiqi@usc.edu). 
This research was supported by funding from the National Science Foundation (NSF) CAREER award \#2240075, the NASA Planetary Science and Technology Through Analog Research (PSTAR) program, Award \# 80NSSC22K1313, and the NASA Lunar Surface Technology Research (LuSTR) program, Award \# 80NSSC24K0127. The authors would like to thank Gabriel Maymon for helping with preliminary data collection.}

\thanks{$^{1}$Shipeng Liu, Meghana Sagare, Shubham Patil, and Feifei Qian are with the Department of Electrical and Computer Engineering, University of Southern California, Los Angeles, CA, USA.
        {\tt\footnotesize shipengl@usc.edu; msagare@uci.edu; svpatil@usc.edu; feifeiqi@usc.edu}
        (\textit{corresponding author: Feifei Qian})}}
        
\maketitle

\begin{abstract}
Deformable substrates such as sand and mud present significant challenges for terrestrial robots due to complex robot-terrain interactions. Inspired by mudskippers, amphibious animals that naturally adjust their tail morphology and movement jointly to navigate such environments, we investigate how tail design and control can jointly enhance flipper-driven locomotion on granular media. Using a bio-inspired robot modeled after the mudskipper, we experimentally compared locomotion performance between idle and actively oscillating tail configurations and found that tail oscillation increased forward speed by 17\% while reducing body drag by 46\%. Shear force measurements revealed that this improvement arises from oscillation-induced fluidization of the substrate, which lowers resistive forces acting on the body. Additionally, tail morphology strongly influenced the oscillation strategy: designs with larger horizontal surface areas leveraged the oscillation-induced reduction in shear resistance more effectively by limiting insertion depth. Based on these findings, we present a design principle to inform tail action selection based on substrate strength and tail morphology. Our results offer new insights into tail design and control for improving robot locomotion on deformable substrates, with implications for agricultural robotics, search and rescue, and environmental exploration.

\end{abstract}

\IEEEpeerreviewmaketitle

\section{Introduction}
  
\IEEEPARstart{R}{ecent} advancements in legged robotic systems have enabled high mobility on a variety of terrains in urban~\cite{lee2024learning, cheng2024extreme, yang2022real} and outdoor environments~\cite{elnoor2024pronav, sorokin2022learning, kumar2021rma}. Despite these improvements, highly deformable substrates like sand~\cite{li2009sensitive,qian2013walking,qian2015principles,hutter2022traversing,choi2023learning,liu2025scout} and mud~\cite{9844167,liu2023adaptation, liu2025adaptive} still present considerable challenges for legged robots. The ability to traverse these complex terrains can empower robots to aid in critical tasks such as search and rescue~\cite{lindqvist2022multimodality}, planetary exploration~\cite{arm2023scientific, liu2024modelling, fulcher2024making, liu2024understanding, jiang2025safe}, and agricultural applications~\cite{liu2023adaptation}. 

To address these challenges, researchers have drawn inspiration from morphology and controls of biological locomotors~\cite{sharpe2015locomotor, mcinroe2016tail,sharpe2013environmental} that excel in these environments to develop robots with enhanced mobility on complex terrains~\cite{maladen2011mechanical, ding2012mechanics, doi:10.1177/0278364911402406, marvi2014sidewinding, astley2015modulation, marvi2014sidewinding, miller2012using, sharpe2015locomotor, qian2015principles}. These biologically-inspired designs and principles have led to robots with improved performance on challenging terrains. For example, by understanding the mechanisms that sea turtles use to move on soft sand, researchers have enabled robots with compliant flipper wrists that generate thrust by solidifying deformable substrates~\cite{mazouchova2013flipper}. Similarly, by understanding how snakes traverse steep sand slopes, researchers have developed snake robots that adjust their body-sand contact length to climb inclined sand~\cite{marvi2014sidewinding}. 

\begin{figure}
    \centering
    \includegraphics[width=1\linewidth]{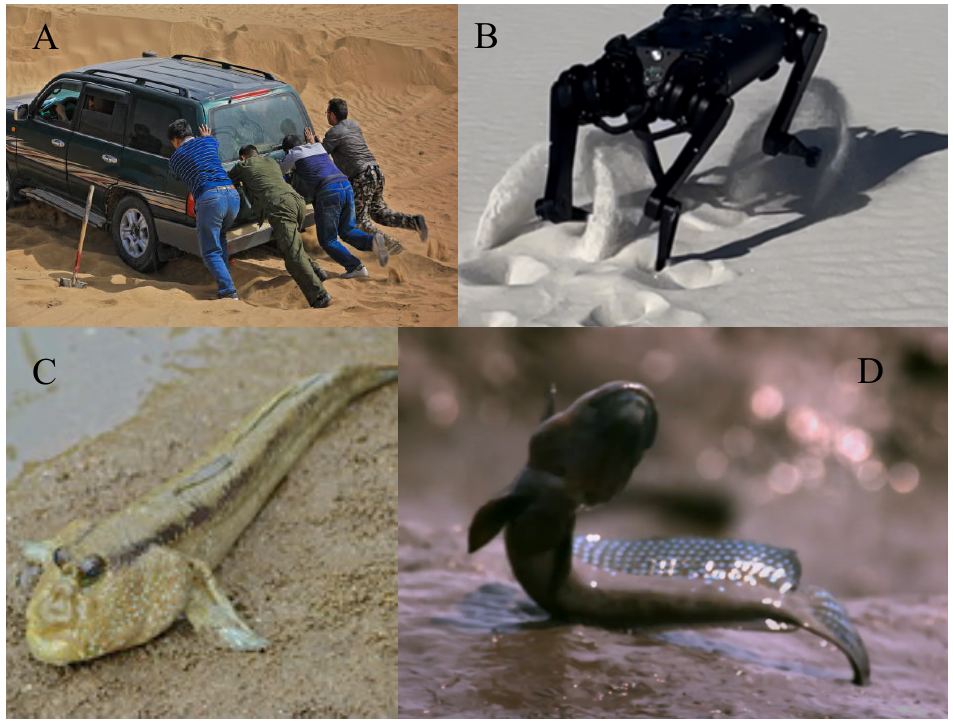}
    \caption{Locomotion challenges on deformable substrates and biologically-inspired solutions. Deformable substrates can cause catastrophic failures for vehicles (A) and robots (B), including sinkage, slippage, or even getting completely stuck. A flipper-driven locomotor, the mudskipper \cite{mudskippertail} (C),  can utilize its tail to effectively move through sand slopes and muddy terrains (D). 
    }
    \label{fig:motivation}
\end{figure}

In this study, we focus specifically on tail-aided locomotion strategies to improve robot performance on deformable substrates. 
Tails are known to play diverse roles in biological and robotic systems~\cite{naylor2022mudskippers,jusufi2008active, libby2012tail, mcinroe2016tail, buckley2023effect, 9444211}, from stabilizing aerial maneuvers during jumps and perching~\cite{libby2012tail, mcinroe2016tail}, to negotiating uneven terrains via compliance and tapping~\cite{soto2022enhancing,buckley2023effect}, to generating additional momentum for slope climbing~\cite{mcinroe2016tail}. However, their potential to mitigate fundamental challenges of locomotion on deformable substrates, particularly sinkage\cite{li2009sensitive,qian2015principles} and slippage\cite{maladen2011undulatory}, remains underexplored.

Mudskippers, amphibious fish that traverse sand and muddy terrains, provide a compelling biological example of tail-enhanced mobility on deformable substrates. They modulate both tail morphology (rolling from a slender profile to flattening into a broad paddle) and tail kinematics (rapid oscillatory strokes versus more static postures) depending on substrate conditions~\cite{naylor2022mudskippers}. This dual adaptation suggests that tail design and control may simultaneously regulate sinkage and drag.

Inspired by this behavior, we investigate how tail morphology and motion can jointly influence robotic locomotion on deformable substrates. Using a robot with interchangeable tails of varying support areas and controllable oscillatory motions, we perform systematic experiments on granular media. Our results show that both increased tail area and tail oscillation motion significantly enhance locomotion speed. To understand this enhancement, we conducted force measurements of substrate penetration and shear resistance under different tail configurations. Our results reveal how tail morphology and motion coupling modulates substrate reaction forces, and leads to principles of tail and control that improve robot mobility on deformable substrates.


\begin{figure*}[bhtp!]
    \centering
    \includegraphics[width=1\linewidth]{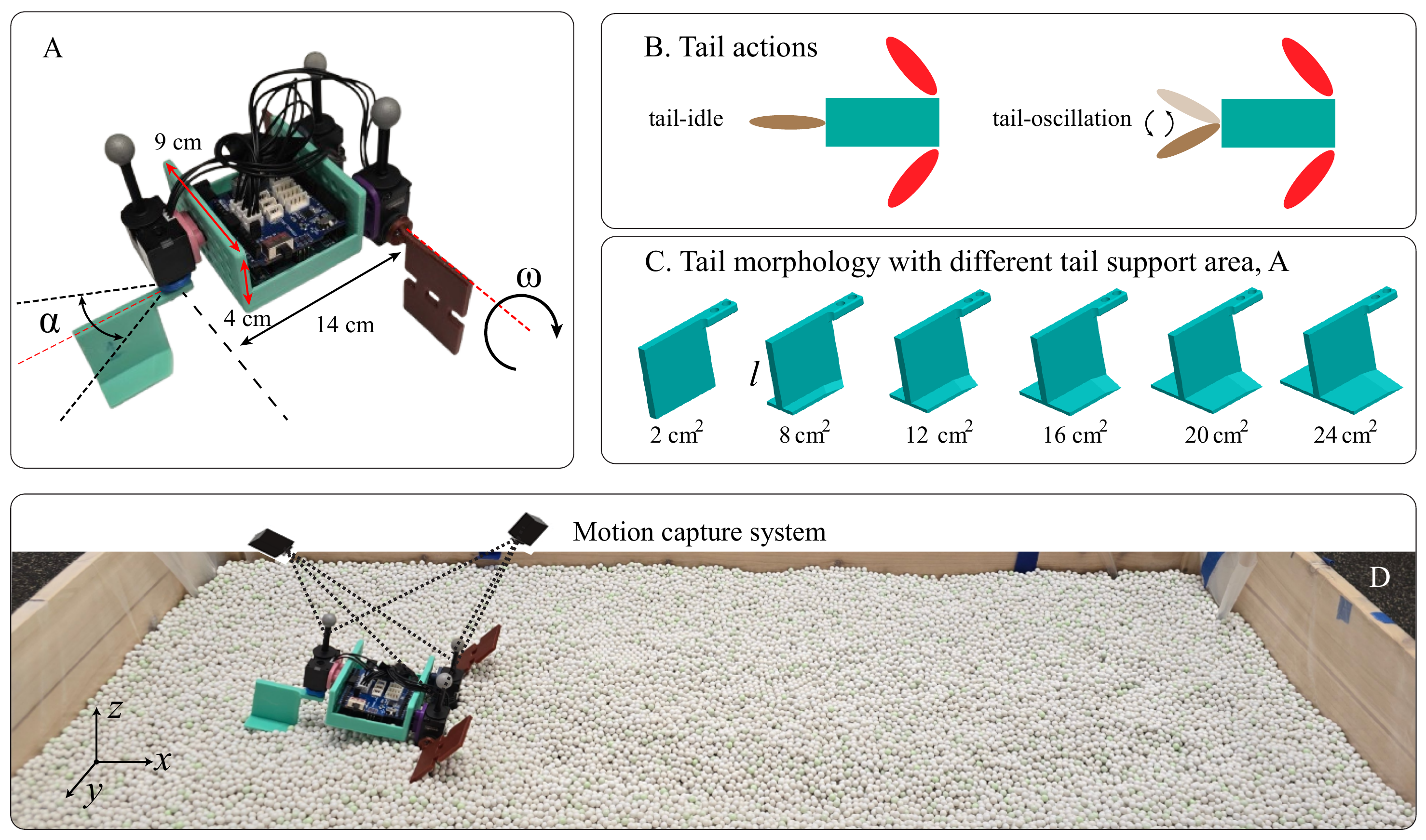}
    \caption{Experimental setup. (A) A mudskipper-inspired robot as a physical model to systematically test the effect of tail oscillation on locomotion performance in granular media. The two front flippers rotate synchronously in the sagittal plane at a constant angular speed, $\omega$, while the tail oscillates horizontally with amplitude $\alpha$ and frequency $f$. (B) Two tail actions were tested: idle tail and oscillating tail. (C) Various tail morphologies were tested, with the tail height, $l$, kept at 40 mm, and the tail support area (\ie the projected surface area perpendicular to the penetration direction), $A$, ranging from 2 cm$^2$ to 24 cm$^2$. (D) Experimental apparatus to assess robot locomotion performance on granular media, with four motion capture cameras tracking the robot's forward and backward movements and a lateral video camera recording the tail-terrain interaction.}
    \label{fig:setup}
\end{figure*}

\section{Materials and Methods} 

\subsection{Robot design and kinematics}\label{sec: robot}

To investigate the effects of tail morphology and movement on locomotion performance, we developed a flipper-driven robot (14 $\times$ 9 $\times$ 4 cm, 135g) modeled after the mudskipper (\textit{Periophthalmus barbarus})~\cite{kawano2013propulsive}. Mudskippers use their two pectoral fins simultaneously in a crutching motion to propel themselves forward on land while also flexing its tail against deformable substrates to generate additional thrust. Their effective locomotion and tail use have inspired the development of flipper-based robots with significantly improved capabilities on granular slopes~\cite{mcinroe2016tail}. In this study, we utilize the mudskipper-inspired robot as a robophysics~\cite{aguilar2016robophysical} model to investigate the potential benefits of tail morphology and oscillation for locomotion on granular media.

To emulate the mudskipper's pectoral fins, two flipper arms  (60 mm $\times$ 40 mm) were 3D-printed from PLA and attached to the front end of the robot body (Fig. \ref{fig:setup}A, B). 
The flippers rotate synchronously at a constant angular speed $\omega$, actuated using two Dynamixel XL-330 servo motors. For all trials in this study, we used a constant frequency of 60 RPM. 

A tail is attached to the rear end of the robot and actuated using an additional Dynamixel XL-330 servo motor, enabling two distinct tail motion modes (Fig. \ref{fig:setup}B): (1) an \textit{idle tail}, where the tail remains stationary at $\alpha = 0\degree$ throughout the trial, and (2) an \textit{oscillating tail}, where the tail oscillates horizontally with amplitude $\alpha = 60\degree$ and frequency $f = 5$ Hz. Both flipper and tail motors are controlled by a microcontroller (Arduino Uno) mounted on the robot body.

To investigate the effect of tail morphology on locomotion performance, we designed interchangeable tails with varying horizontal support areas (Fig. \ref{fig:setup}C). All tail designs maintain a consistent vertical height $l = 40$ mm, while the horizontal support area, $A$, ranges from 2 cm$^2$ to 24 cm$^2$ (corresponding to weights between 8-11g). This design allows systematic investigation of how tail support area affects substrate interaction and locomotion performance.

\subsection{Locomotion experiments}\label{sec:setup}

Locomotion experiments were conducted in a 115 cm × 54 cm × 12 cm trackway (Fig.~\ref{fig:setup}D). The substrate consisted of \href{https://www.evike.com/products/24127/Matrix-Match-Grade-6mm-Airsoft-BBs-Color-.25g-10000-Rounds-White/}{6 mm spherical plastic particles} (Matrix Tactical Systems) as a model granular medium. The 6 mm particles were chosen as they behave rheologically similarly to natural sand and soil~\cite{doi:10.1126/science.1172490,li2013terradynamics}, while the simpler geometry facilitated systematic control and preparation of the substrate for laboratory experiments. 

To investigate how the tail affects flipper-driven locomotion performance on granular media, we performed two sets of locomotion experiments with systematically varied tail morphologies and motions. The first experiment investigated the effect of tail oscillation, comparing oscillating tail performance against idle tail as a control. Additionally, inspired by the mudskipper's tail-morphing behavior of rotating and flattening its tail on soft terrain, we investigated the effect of tail support area by comparing robot speed between a ``vertical'' tail (with smaller supporting area) and a ``flat'' tail (with larger supporting area). For both experiments, three trials were performed for each configuration. After each trial, the terrain surface was reset manually using a scraper.

To capture robot locomotion performance and robot-substrate interactions, four Optitrack Prime 13W motion capture cameras were installed above the trackway to monitor the robot's position in the world frame (x - fore-aft; y - lateral; z - vertical) at 120 frames per second. An additional side-view camera was used to observe the robot's height and flipper interactions with the granular medium.

\section{Results and Discussion}

\subsection{Effect of tail morphology and motion on robot speed}\label{sec:speed}

\begin{figure}[tbhp!]
    \centering
    \includegraphics[width=1\linewidth]{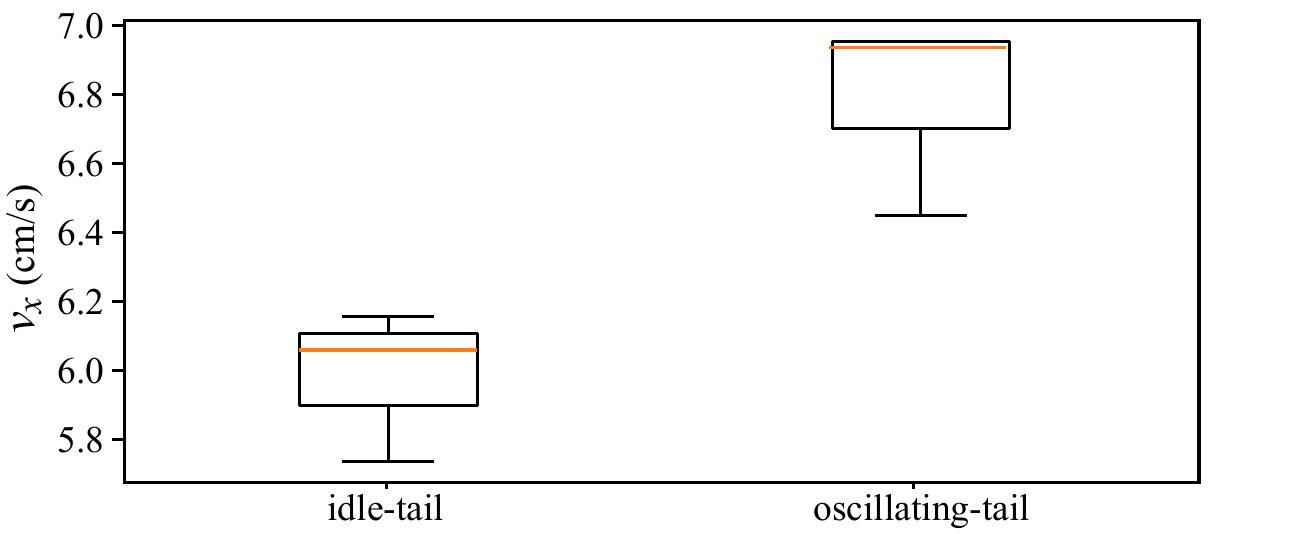}
    \caption{Comparison of average forward speed ($v_x$) of the robot between idle and oscillating tails. Results are shown for tail size $ A=$16 cm$^2$.}
    \label{fig:overall_results}
\end{figure}

The robot's average forward speed, \(v_x\), showed a clear difference between idle and oscillating tail configurations (Fig. \ref{fig:overall_results}).
With an oscillating tail, the robot achieved a speed of 6.9 $\pm$ 0.23 cm/s, representing a 17\% improvement over the idle tail condition (5.9 $\pm$ 0.18 cm/s).  

\begin{figure}[]
    \centering
    \includegraphics[width=0.95\linewidth]{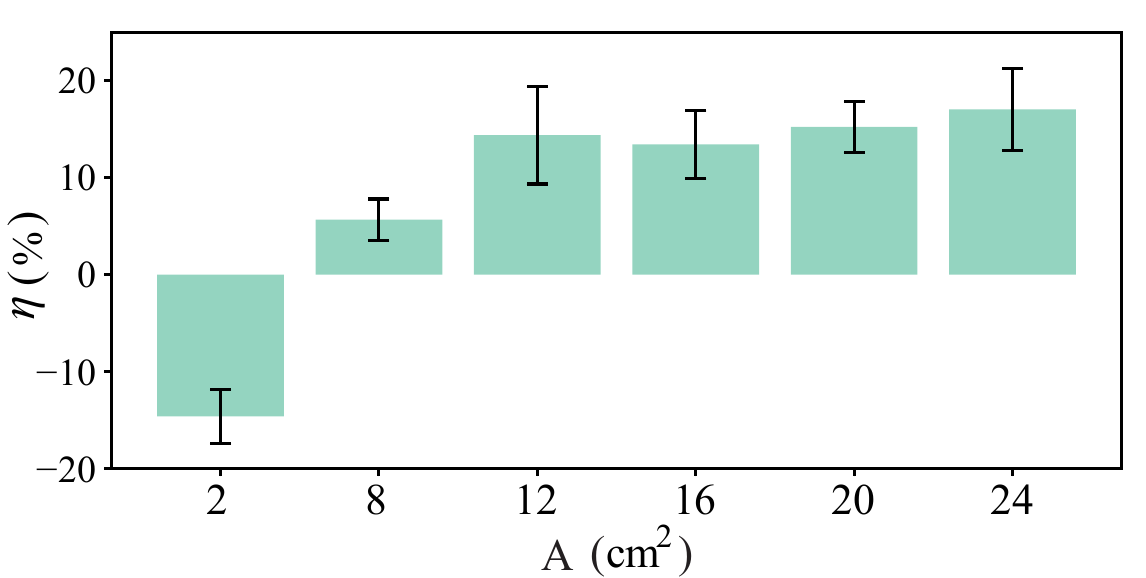}
    \caption{
    Percent improvement in averaged forward speed ($\eta = (v_o - v_i) / v_i$) with an oscillating tail compared to the idle tail, for different tail support surface areas, $A$. }
    \label{fig:basesize}
\end{figure}

Interestingly, tail oscillation did not universally improve robot performance. Speed measurements across different tail sizes, \(A\), reveal that larger tails (8-24 cm\(^2\)) benefited most from oscillation, with gains of \(\eta = 6\%-20\%\) compared to idle tails (Fig. \ref{fig:basesize}). 
In contrast, smaller tail base areas (\(A < 8\) cm\(^2\)) showed diminishing benefits, with the smallest tail (\(A = 2\) cm\(^2\)) even reducing speed under oscillation. 

These results demonstrate that the effectiveness of tail oscillation depends critically on tail morphology. This coupling introduces a co-design challenge where both tail shape and movement must be jointly optimized to enhance mobility. 
To uncover the physical mechanism underlying this coupling and to inform systematic design guidelines, we next analyze the substrate reaction forces through direct measurements and modeling. 


\subsection{Shear force analysis reveals that tail oscillation enhances robot speed via sand fluidization}\label{sec:oscillation-effect}

To investigate the mechanism behind the speed improvement from tail oscillation, we measured the horizontal force exerted on the robot's body under both idle and oscillating tail configurations. 

\begin{figure}[bhtp!]
    \centering
    \includegraphics[width=1\linewidth]{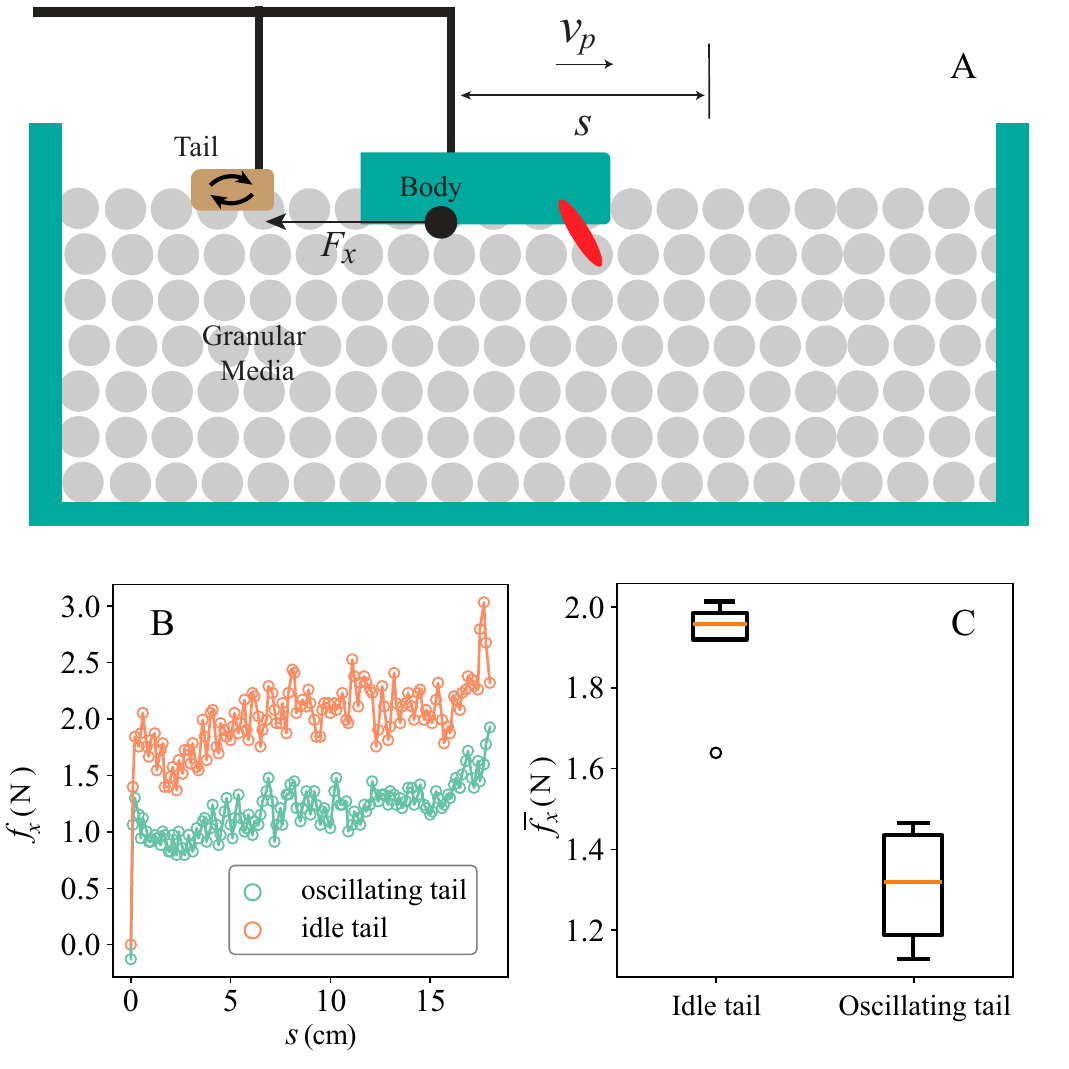}
    \caption{Shear force experiments. (A) Schematic of the force data collection setup. The robot body is dragged horizontally through the granular media over a total distance of 18 cm with a constant velocity of 2 cm/s, at a fixed insertion depth of 1 cm. The shear resistance force exerted on the robot body, ${f_x}$, was measured using a force sensor. (B) Shear resistance force measured from both the oscillating and idle tail conditions, plotted against the shear distance, $s$. (C) Comparison of the averaged shear force, $\bar{f_x}$, between the idle tail and oscillating tail. $\bar{f_x}$ was computed as the averaged shear resistive force, ${f_x}$, between $s$ = 2.5 cm and $s$ = 17.5 cm.}
    \label{fig:shear}
\end{figure}

To isolate the shear component of resistance relevant to forward locomotion, the robot was mounted on an actuated horizontal linear stage (Fig. \ref{fig:shear}A), and dragged horizontally through the granular medium at a constant speed of 2 cm/s and an insertion depth of 1 cm. A force sensor (DYMH-103) was installed between the robot and the linear stage to measure shear resistive forces exerted on the body (i.e., body drag). 
Unlike free-running trials, this setup constrained pitch and roll, enabling direct comparison between idle and oscillating tails. To further minimize force perturbations from tail-induced wiggling, the tail ($A = 16$ cm$^2$) was mounted separately at the same distance behind the body as in locomotion experiments.

Two tail modes were tested: (1) idle, and (2) oscillating at $f = 5$ Hz with amplitude $\alpha = 60\degree$. Each mode was repeated three times. Results showed that after the initial pull of the linear actuator, the robot body consistently experienced lower resistive forces with an oscillating tail compared to an idle tail (Fig. \ref{fig:shear}B). As shown in Fig.~\ref{fig:shear}C, tail oscillation reduced the average shear resistive force on the robot body by 46\% relative to the idle condition. This drag reduction provides a mechanistic explanation for the increase in locomotion speed observed in free-running experiments: with lower resistance opposing forward motion, the robot can traverse the granular medium more efficiently.

This finding aligns with prior studies in granular physics, which show that mechanical vibrations can fluidize granular media and reduce resistive forces. For example, controlled vibrations were found to reduce shear stress in dry sand~\cite{xie2022experimental,jiang2022vibro}, and head wiggling decreases drag by up to 45\% during burrowing motion through wet sand~\cite{sharpe2015controlled}. 



\subsection{Large tail size enhances tail oscillation strategy by reducing sinkage}\label{sec:size-effect}


\begin{figure*}[hbtp!]
    \centering
    \includegraphics[width=1\linewidth]{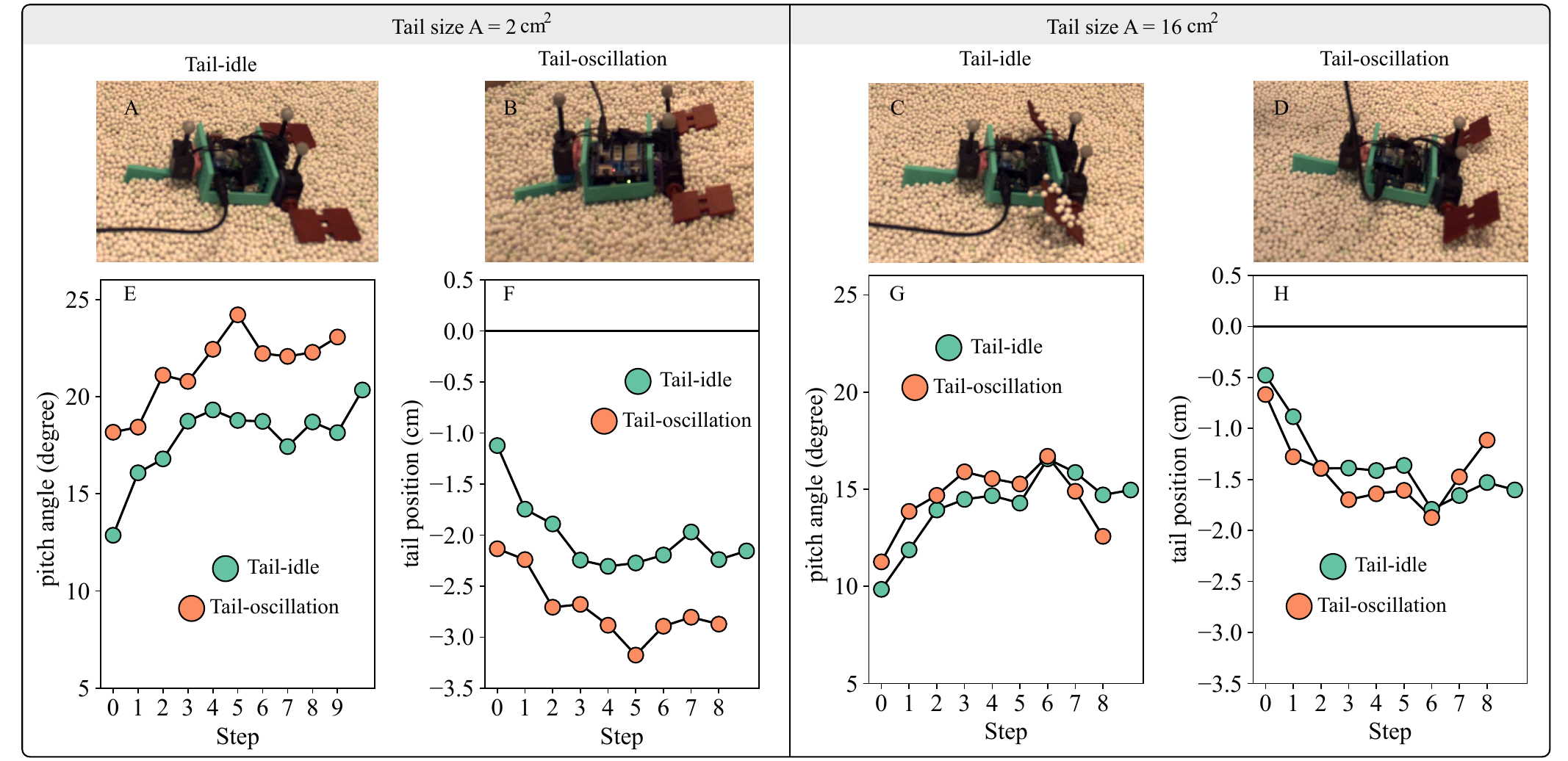}
    \caption{Comparison of representative locomotion trials of tail-idle and tail-oscillation with different tail sizes, 2 cm$^2$, and 16 cm$^2$. (A-D) Side view pictures during locomotion. (E)(G) Robot pitch angle versus step. The positive pitch angle indicates upward body tilt.  (F)(H) Robot tail z-position versus step. The black horizontal line ($y$ = 0) represents terrain surface height. Green and orange markers represent the idle and oscillating tails, respectively.}
    \label{fig:video_insertiondepth}
\end{figure*}


To understand why oscillation improved performance only with large tail sizes, we examined side-view videos (Fig. \ref{fig:video_insertiondepth}A-D) and tracked robot pose (Fig. \ref{fig:video_insertiondepth}E-H) for trials with small ($A = 2$ cm$^2$) and large ($A = 16$ cm$^2$) tails. 

With the small tail ($A = 2$ cm$^2$), oscillation induced noticeably greater body pitch (Fig. \ref{fig:video_insertiondepth}B, E) compared to idle tail (Fig. \ref{fig:video_insertiondepth}A, E). In contrast, with the larger tail ($A = 16$ cm$^2$), the robot's pitch angle remained nearly identical between the two tail configurations (Fig. \ref{fig:video_insertiondepth}G). Tracking data further revealed that this increased pitch stemmed from greater insertion depth of the robot's rear section. For the small tail ($A$ = 2 cm$^2$), the rear section sank deeper into the granular media under tail oscillation (Fig. \ref{fig:video_insertiondepth}F), whereas with the larger tail ($A$ = 16 cm$^2$) the insertion depth remained similar between idle and oscillating tails (Fig. \ref{fig:video_insertiondepth}H).

These observations suggest a tradeoff between two opposing effects of oscillation: (i) fluidization of the substrate reduces shear strength and drag, but (ii) the softened substrate also lowers normal support, increasing sinkage. Since resistive force in granular media scales with both the projected area and insertion depth of the intruder~\cite{albert1999slow}, the net effect depends on tail size. For smaller tails, the increased shear force on the body due to the increased sinkage outweighs the reduced shear force from the fluidized sand, diminishing or even reversing performance gains. In contrast, a larger tail base provides sufficient vertical support to limit sinkage, allowing oscillation-driven fluidization to dominate and enhance locomotion.
 
\subsection{Co-design principle: jointly adapting tail morphology and motion to improve mobility in sand}\label{sec:model}

 To test the hypothesized tradeoff and guide the optimization of tail size and oscillation motion, we developed a physics-based model to quantify the trade-offs between terrain fluidization-induced drag reduction and sinkage. The model provides two key capabilities: (1) predicting sinkage as a function of tail size, to capture the observed effect of morphology, and (2) predicting body drag under idle and oscillating conditions, to identify optimal combinations of tail morphology and motion. 

\paragraph{Modeling sinkage to capture tail morphology effect}

\begin{figure}[bhtp!]
    \centering
    \includegraphics[width=1\linewidth]{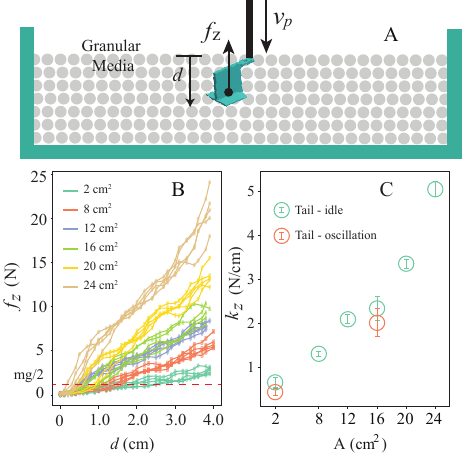}
    \caption{Penetration force experiment. (A) Experiment setup. The tail is lowered into the medium to an insertion depth, $d$, with a constant speed, $v_p$. The penetration resistance force, $f_z$, is measured using a load cell. (B) Experimentally-measured penetration resistance force for tail base areas ranging from 2 cm$^2$ to 24 cm$^2$. Color represents tail size. The red dashed line represents the applied penetration force during locomotion, approximated as half of the robot’s weight. The intersections between the applied load and the penetration resistive force  curves yield the predicted sinkage. (C) Penetration resistance, $k_z$, for each tail base area, $A$. The green and orange markers represent idle and oscillating tails, respectively.}
    \label{fig:penetration}
\end{figure}

The model assumes that the penetration resistive force, $f_z$, balances the weight of the robot's rear section, \(\frac{1}{2}mg\) (dashed red line in Fig. \ref{fig:penetration}B). This balance yields an estimation of penetration depth, $d$, for substrates with a given normal strength. 

To experimentally measure substrate strength, we performed vertical penetration tests (Fig.~\ref{fig:penetration}A). Using a linear actuator (Heechoo), each tail was vertically lowered into the same granular medium used in locomotion trials to a depth of 4 cm at 2 cm/s. The low speed was chosen because granular resistive forces are largely rate-independent below impact speeds of 20–30 cm/s~\cite{albert1999slow}. A force sensor (DYMH-103) was used to measure the penetration resistive forces exerted on the tail. To evaluate how different tail sizes affect sinkage depth, six different base areas were tested, with five trials per condition. 

For all trials, the penetration force, \( f_z \), increased linearly with the insertion depth, \( d \) (Fig. \ref{fig:penetration}B). This is consistent with previous granular physics literature~\cite{albert1999slow,katsuragi2007unified}. The slope of force per depth, \( k_z \), referred to as the penetration resistance, captures the normal strength of the deformable terrain in granular physics~\cite{nedderman1992statics,qian2015principles}. Using the measured values of $k_z$, the model estimates the sinkage for each tail size. 

\begin{figure}[bhtp!]
    \centering
    \includegraphics[width=1\linewidth]{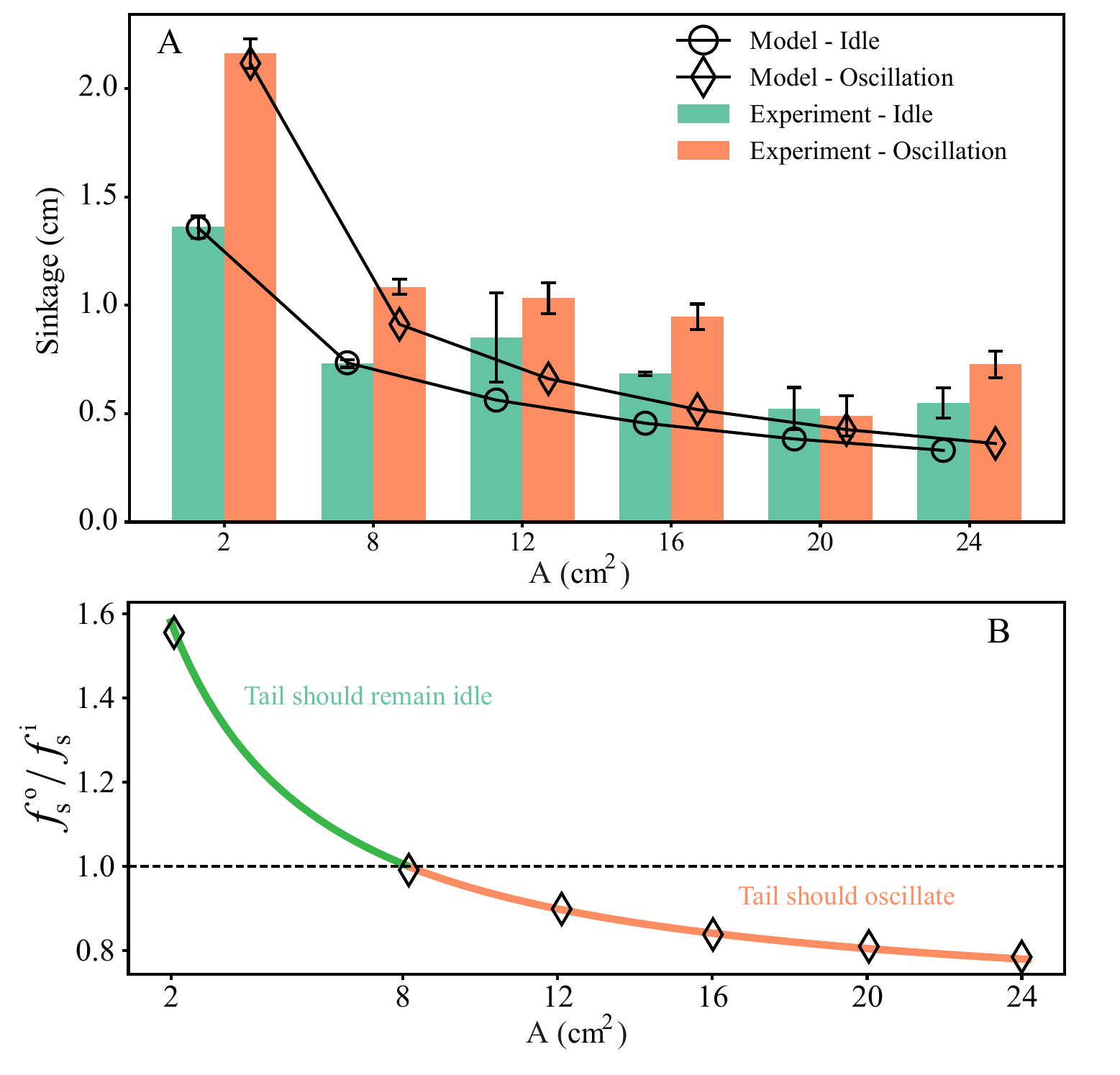}
    \caption{Model-informed co-design principle. (A) Model-predicted sinkage for idle (black circles) and oscillating (black diamonds) tails, compared against experimentally-measured sinkage for idle (green bars) and oscillating (orange bars) tails, averaged from the first three steps of locomotion trials. $A$ is the tail size. (B) Model-predicted ratio of body drag, for an oscillating tail to an idle tail. The green segment represents the range of size where an idle tail is beneficial, where the body resistance force during tail oscillation, \( f_s^o \), is greater than that during idle, \( f_s^i \). Conversely, the orange segment, where \( f_s^i \) exceeds \( f_s^o \), represents the range where tail oscillation could improve mobility.}
    \label{fig:overall_depth}
\end{figure}

The model-predicted sinkage captured the trend observed from the locomotion trails (Fig. \ref{fig:overall_depth}A), and 
explained the size-dependent effectiveness of the oscillation strategy. As discussed in Sec. \ref{sec:oscillation-effect}, tail oscillation locally fluidizes the granular medium, reducing penetration resistance. Notably, the amount of reduction was comparable for small ($A$ = 2 cm$^2$) and large ($A$ = 16 cm$^2$) tail sizes (Fig. \ref{fig:penetration}C). Consequently, smaller tails experienced disproportionately greater sinkage under oscillation (Fig.~\ref{fig:overall_depth}A), diminishing the locomotion benefits observed in experiments.

\paragraph{Modeling body drag to co-optimizing tail morphology and action}

Using the modeled sinkage, we further computed the body shear resistance force for oscillating and idle tails (Fig. \ref{fig:overall_depth}B, $f_s^o$ and $f_s^i$, respectively).  The shear resistance force on a segment at insertion depth $d$ scales with the segment's projected area and its depth~\cite{albert2001stick}. Thus, the total shear resistance on a tail can be computed by integrating the segmental forces~\cite{li2013terradynamics}. For each tail size, the net change in body resistive forces arises from two competing effects: the reduced substrate strength due to localized fluidization, and the increased insertion depth. 

The model prediction shows a monotonic decrease of oscillation-induced body drag as tail size increases. For small tails ($A < 8$ cm$^2$), the increased insertion depth dominates, causing oscillation to increase net body resistance (Fig. \ref{fig:overall_depth}B, green region). In this case, an idle tail is more beneficial as it minimizes sinkage. In contrast, for sufficiently large tails ($A > 8$ cm$^2$), oscillation reduces body resistance while maintaining a nearly constant insertion depth (Fig. \ref{fig:overall_depth}B, orange region), making the oscillation strategy more advantageous.

Together, these results establish a co-design principle: effective tail use on deformable terrain requires simultaneous consideration of morphology and motion. Large, flat tails provide the penetration support needed to counteract local fluidization, while oscillation reduces shear resistance and jointly enables enhanced forward speed. This principle may also shed light on animal strategies --- for instance, why mudskippers rotate their tails to engage the flattened side when moving on deformable substrates.

\section{Conclusion}

This study demonstrates that tail movement and morphology jointly influence robot locomotion performance in deformable granular substrates. When the tail support area is sufficient ($A \geq 8$ cm$^2$), oscillation locally fluidizes the medium, reducing body drag by 46\% and enabling a 17\% speed increase. In contrast, smaller tails ($A < 8$ cm$^2$) increase sinkage and reduce speed, highlighting the need to optimize tail shape and motion together.

Our key contribution is the establishment of a tail co-design principle for bio-inspired locomotion on deformable substrates: oscillation fluidizes the substrate but also promotes sinkage, and only adequate tail size allows the benefits of oscillation to outweigh the drawbacks. This finding reframes tail design as an integrated system problem, showing how morphology and motion must be optimized in tandem to enhance locomotion performance.

Our results lay the foundation for future research to explore additional strategies to improve mobility in challenging deformable media, such as tail oscillation frequency modulation or tail shape modification for more efficient drag reduction across different deformable terrains. These principles are crucial for enabling next-generation bio-inspired robots with the ability to traverse natural, complex terrains such as sand and mud, and to assist in applications such as disaster recovery, planetary exploration, and agricultural operations.


\bibliographystyle{IEEEtrans}
\bibliography{main}

\end{document}